# Implementation of Auto Monitoring and Short-Message-Service System via GSM Modem

Akilan Thangarajah[1], Buddhapala Wongkaew[2], Mongkol Ekpanyapong[3]

[1]*Department of Electrical and Computer Engineering, University of Windsor, Canada*
[2,3] *Industrial Systems Engineering, Asian Institute of Technology, Thailand*
*Email:* [1]*thangara@uwindsor.ca,* [2]*Buddhapala.Wongkaew@ait.ac.th,* [3]*mongkol@ait.asia*

***Abstract-*** **Auto-Monitoring and Short-Messaging-Service System is a real-time monitoring system for any critical operational environments. It detects an undesired event occurring in the environment, generates an alert with detailed message and sends it to the user to prevent hazards. This system employs a Friendly ARM as main controller while, sensors and terminals to interact with the real world. A GSM network is utilized to bridge the communication between monitoring system and user. This paper presents details of prototyping the system.**

## I. INTRODUCTION

Technological advancement in sensors, microcontrollers, and mobile communication opened vast opportunity for various applications. Among them real-time monitoring has become one of the most demanding applications with the integration of available technology. Although, there are several real-time monitoring systems were developed in the field of healthcare, production line management, transportation as described in [1] - [7] still there are rooms for improvements, new implementations, and experiments. Thus, we prototyped a surveillance system for real-time operational environments named Auto Monitoring and Short Message Service System via GSM module, we call it AMSMS. This system monitors an operational environment 24/7 with the help of multiple sensing devices. Whenever it detects an undesired event happening in the environment it generates an alert with detailed message and sends it to the officials or person in charge to prevent hazards. This system, technically, is cost effective when considering commercially available real-time machine monitoring/controlling systems. It is also feasible to commission this AMSMS wheresoever there is a doubt in safety of a manual inspection or monitoring staff in a critical operational environment. Although the system is novel, it is at the experimental stage and looking for relevant improvement in the future. The following assumption is made for prototyping the AMSMS.

An automatic flagging system must be implemented to monitor the toxic level of gas, level of temperature, and unauthorized access of an interested operational environment which is in a remote location. The flags must be transmitted to the person in charge in daily and emergency basis. The content of the message must include the status of the monitored environment with location i.e. coordinates or index of the monitored environment by default. Based on this assumed requirements, the Auto Monitoring and Short Messaging Service System via GSM modem is prototyped with integration of sensors to capture the interested particles in the environment, module of a Global System for Mobile (GSM) Communications to send flags, and a micro-controller unit (MCU) to interface the devices, to monitor the operations, and to control the entire system. The rest of this paper is organized as follows. Section 2 describes the system architecture and design, Section 3 emphasizes the implementation techniques, Section 4 provides the system testing and result, and Section 5 concludes the paper.

## II. SYSTEM DESIGN AND ARCHITECTURE

This section describes the system design and architecture by block diagrams, and flow chart.

*Hardware Architecture*

Abstract level of hardware architecture of the system is described in Figure - 1. The sensor module consists of temperature sensor, light dependent register (LDR), and variable resistors (acting as smoke sensor). The communication module is a GSM modem. The terminal belongs to a respective operating system or any compatible third party application to execute system commands, and MCU is the brain of the entire system. Detailed hardware integration is visualized in Figure - 2. List of employed hardware components and their specifications are given in Tables 1 - 4.

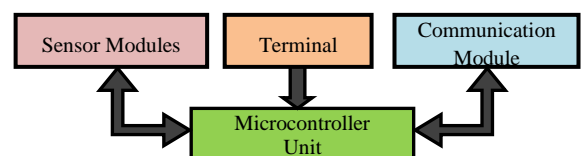

Figure 1: Block Diagram: Hardware Architecture.

*Implementation of Auto Monitoring and Short-Message-Service System via GSM Modem*

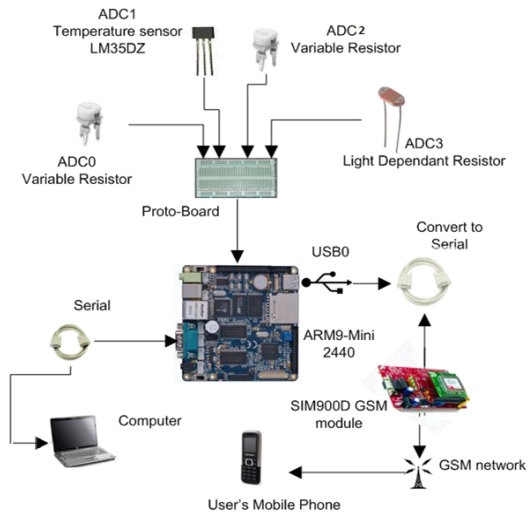

Figure 2: Interface diagram of AMSMS [8] - [10].

This system employs a Mini2440 Friendly ARM-9 which is a practical low-cost embedded system development board as main controller, efficacious and robust for wide range of Applications [2] – [5]. A SIM900D GSM/GPRS module (refer Figure – 3.a) is used as a backbone of the auto-messaging functionality. The GSM was introduced by the European Telecommunications Standards Institute (ETSI) as replacement of the first generation (1G) analog cellular networks. The GSM standard originally described a digital, circuit switched network optimized for full duplex voice telephony. However, the standard was expanded over time to include first circuit switched data transport, then packet data transport via GPRS (General Packet Radio services) as well [11]. Such data transmission feature is the key for this AMSMS.

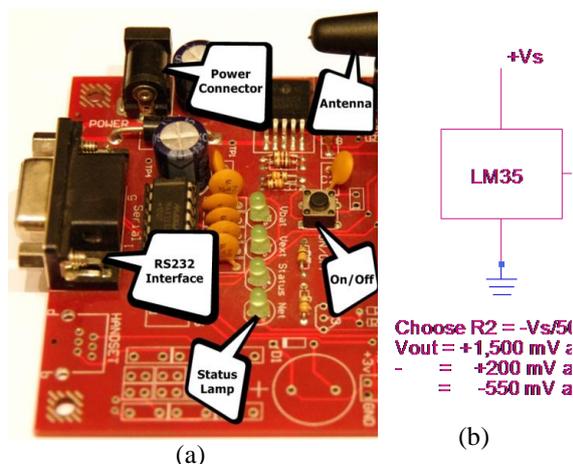

Figure 3: Modules: (a) SIM900D GSM Module, (b) LM35DZ [10].

Sensors are one of the most fundamental components of any control systems. There are various types of them which respond to changes of a physical phenomenon such as heat, light, sound, pressure, magnetism, or a particular motion and generate a resulting impulse. In this AMSMS, an LM35DZ series temperature sensor is circuited to capture the temperature of an operational environment. The output voltage of this circuit is nearly proportional to the temperature experienced by the sensor in the concerned environment. LM35 series is also capable to calibrate over linear temperature sensors in Kelvin. It, also, has low output impedance, linear output and precise inherent calibration [10]. Figure – 3.b shows the circuit diagram of the temperature sensor LM35DZ where, the readings are in full scale. However, AMSMS utilizes the basic centigrade calibration (0 - 100 $^{o}$C). Similarly, the system uses an LDR for detecting trespassing. LDRs can be used as light sensors. Normally, resistance of an LDR is very high. Sometimes it goes as high as 1MΩ, but when they are illuminated with light, the resistance drops dramatically. Here, the resistance of the employed LDR ranges 200Ω – 200 kΩ.

Table 1: Hardware List of AMSMS.

| MCU | Mini2440 S3C2440 ARM9 |
|---|---|
| Communication Module | SIM900D evaluation board |
| Sensors | Light Detecting Resister (LDR) Variable resistance Temperature sensor (LM35DZ) |

Table 2: Specification- Mini2440 FriendlyARM [8].

| Processor | Samsung S3C2440A (400MHz) |
|---|---|
| Interfaces | 3 serial ports, 1 USB host (B-type) SD card storage |
| Expansion | 134-pin GPIO |
| OS | Linux 2.6.29, Windows CE .NET 5.0 |
| Others | 4 User LEDs, 1 buzzer PWM control 1 adjustable resistor for A/D test |

Table 3: Specification- SIM900D GSM Module.

| Control: via AT commands | Power consumption: 1.0mA(sleep mode) |
|---|---|
| Supply voltage range: 3.2 - 4.8V | Operation temperature: -40 to +85 $^{o}$C |
| SMS: via GSM/GPRS Modes: Text and PDU mode | Interfaces: Serial interface, Antenna pad |

Table 4: Specification- LM35DZ Sensor [10].

| Calibrated directly in $^{o}$C | Linear +10:0mV/C scale factor |
|---|---|
| Rated for full -55 to 150C range | Low impedance output 0.1 for 1mA load. |
| Operates from 4 to 30 volts | Less than 60A current drain |

## *Software Architecture*

The software architecture of the system is shown in Figure - 4. This system employs Linux OS (kernel Ver.2.6.32-33). The application interfaces system developer(s) and OS. We utilized Linux tools such as *gcc, gedit, shell terminal,* and the *GTK terminal* for developing, debugging, and compiling the system application. Meanwhile, the application interface also facilitates the developer(s) to program the MCU.

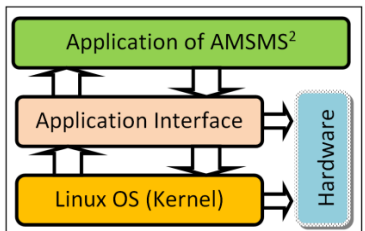

Figure 4: Software Architecture of the System

## III. IMPLEMENTATION

We took advantage of "divide and conquer" methodology for the successful implementation of the system. Initially, the main components of the system were individually programmed and tested. For example, GSM module was directly connected to a personal computer (PC) through serial communication (COM) and programmed from a terminal (in this case the GNOME Linux terminal called *GTK* was utilized). Similarly, other devices, also, were interfaced with MCU, programmed, and experimented to verify their electrical-electronic characteristics and functionality. Once functionality of all the individual components was tested they were integrated one by one to provide the overall system functionality. Software application of the system was implemented based on the operational flowchart shown in Figure - 5.

The main function of the application is detecting an abnormal condition in the interested environment with the help of hardware components, translating them into user understandable/desirable formats, recording them and then sending a message to warn the user via GSM communication network if necessary. Thus, the entire Software Architecture of the system can be divided into four main sections: *(i). Sensors reading, (ii). Data processing and translation, (iii). Message generation and Status update* by printing and writing data report, and *(iv) Communication between MCU and GSM module*. Each of them is elaborated in the following subsections.

## *Sensor Reading*

The program starts with initializing variables and setting necessary GPIOs required for the process. Consequently, the program will read ADC 0 to 3, store the raw values, and convert them into equivalence voltage one at a time. The raw data is in 10-bit which contains a reading from 0 to 1023. This 10-bit value represents the voltage at the respective ADC port. The voltage ranges from 0 mV to the maximum of 3300 mV. The conversion of the 10-bit data is given in the equation 1. Note that unit of the voltage is mV instead of Volt (V). It is to get rid of floating point operations in the MCU and to deal with integer operations to speed up the process (It was also found that printing and processing float values have issues with MCU).

$$Voltage = 3300 \times \frac{<10-bitvalue>}{1023} mV \quad (1)$$

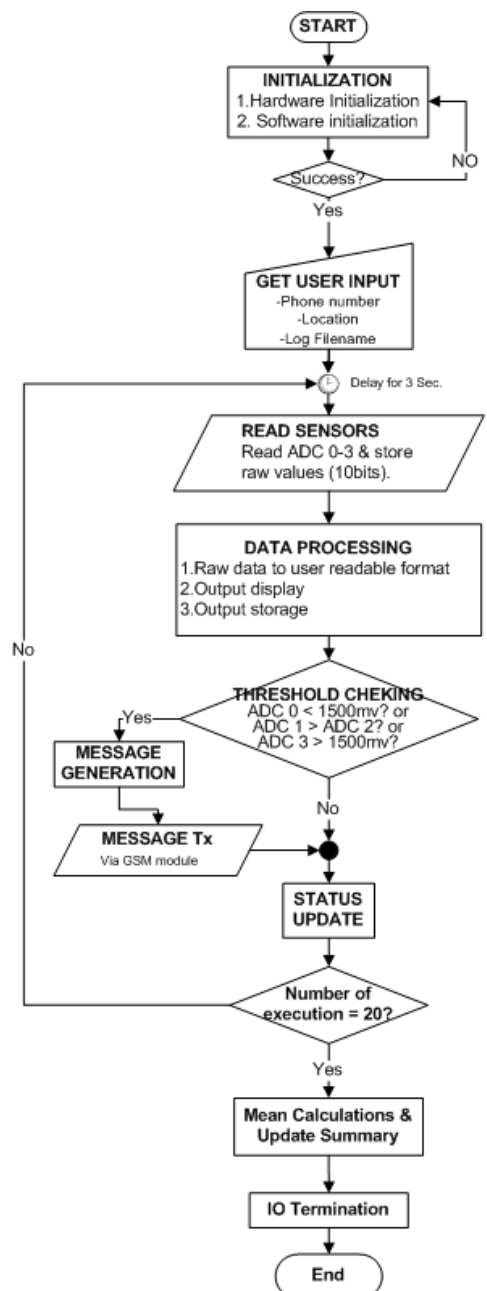

Figure 5: Flowchart of the Overall System Operation.

*Implementation of Auto Monitoring and Short-Message-Service System via GSM Modem*

## Data Processing

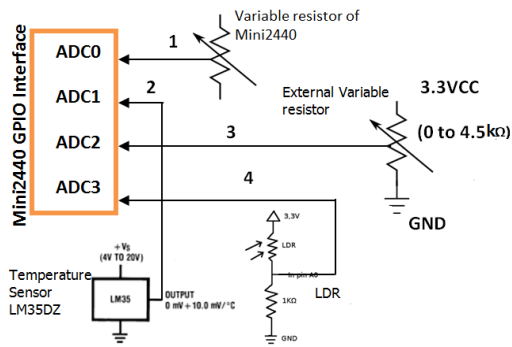

Figure 6: Sensing Interface with Mini2440-ARM9.

As shown in Figure – 6, the system uses four ADCs numbered 0 to 3, to receive input from two variable resisters, a temperature sensor, and an LDR respectively. Representation of each ADC port is given below.

*ADC 0: Smoke detector* - The built in variable resistor on Mini2440 board is used in place of a smoke sensor (which could be a set of Infrared LED and LDR). In general, reading of a smoke sensor is normally a voltage that varies depending how much the LDR is exposed to the infrared light. However, the detection of the smoke should be either detected or undetected. To achieve the two states, a threshold value must be given; in this case, the threshold is set to 1500mV.

Thus, if $output\ voltage < 1500mV$ then smoke is detected else undetected.

*ADC 1: Temperature sensor* - A temperature sensor LM35DZ is connected to ADC 1. The temperature will be read between 2 and 100° Celsius (using basic centigrade mode). The voltage starts with 20mV and additional 10mV per Celsius. The temperature can be converted from Voltage to Celsius by the equation 2.

$$Temperature = \frac{Voltage}{10}\ ^\circ C \quad (2)$$

*ADC 2: Temperature Threshold reference* - A reliable sensor should have an adjustable threshold so that the sensor can adapt to different environment and usage. As keep that in mind, ADC2 is connected to a variable resistor. The voltage appears across this resistor is treated as a reference or threshold temperature. Since it is a normal variable resistor, the value will vary from 0 to 3300mV. This voltage will be converted into temperature reference in the range of 0 to 100 Celsius using equation 3.

$$Temperature\ Ref. = \frac{Voltage \times 100}{1023}\ ^\circ C \quad (3)$$

In equation 3, the voltage can be calculated from the raw 10-bit data by using equation 1, whereby the temperature reference is a parametric quantity i.e. the threshold value to determine whether the operational environment under normal condition or overheated.

Thus, if $temprature\ reading > ref.\ temperature$ then over heating is detected else undetected.

*ADC 3: Trespasser Detecting Sensor* - An LDR is connected to ADC 3. Its output (voltage) varies from about 200mV to 3200mV depending on level of light exposure to the sensor. The light source in this application is either an electric bulb or sunlight. The system utilizes the characteristic of LDR towards light intensity. Its resistance is inversely proportional to the light intensity i.e. when there is a higher light intensity exposed to the LDR, its resistance becomes lower and vice versa. Thus, when a person passes across the LDR, he/she blocks the light source. It causes lack of light exposure to the LDR, and then resistance of LDR becomes high.

Consequently, the LDR yields higher voltage reading as ohm's law, $V=I \times R$. It is found that the threshold output voltages of the LDR are 2000mV and 1500mV for a person blocking the sensor from a light source in a room and from the Sun light respectively. Thus, for the case of trespassing scenario with the present of a room-light source if $voltage > 2000mV$ then a trespassing is detected else undetected.

## Message Generation and Status Updates

After reading and interpreting the raw data based on the expressions 1 - 3, the outcome will be displayed on the terminal for user observation or debugging. The same set of data will be logged in a file so that the user has a chance to backup or trace the detail of the whole day monitoring. The data set includes voltage readings of *ADC 0 - 3*, translation of the voltage reading for *smoke detection, Temperature overheating, Temperature reference, Trespassing*, and a *warning* with status update. The three undesirable conditions will be represented by an 8-bit variable named *status* as shown below.

$$When\ (SmokeDetected)\ Bit\ 0 = 1\ else\ 0$$
$$When\ (OverheatingDetected)\ Bit\ 1 = 1\ else\ 0 \quad (4)$$
$$When\ (TrespassingDetected)\ Bit\ 2 = 1\ else\ 0$$

By having these 3-bit there will be 8 possible statuses can be represented. For example, 000 or 0 means there is no undesirable condition detected. Similarly, 101 or 5 means smoke and trespassing are detected. After these detections the outcome will be stored, a warning message generated based on the status. The warning message will include the location, time stamp, temperature, and summary of the status report. The message is, then, sent via GSM module.

## Communication with GSM Module

The AMSMS2 employs SIM900D GSM module. Generally, command to a GSM module can be sent through any serial communication applications such as *Windows HyperTerminal, Linux GTKterm.* or a third party serial communication applications. When a valid command is sent to GSM module, it will attempt to do the task and acknowledge to the command. In this system, communication between the MCU and GSM module is not established via the traditional serial communication RS232 but through USB connection. The USB port connection is converted to RS232 before connecting to the module (refer Figure – 2). The program utilizes *system( )* call to request Linux OS to execute *echo* commands to send instructions to the GSM module through terminal. The communication with the GSM module is directly dealt with the respective device file */dev/USB0*. The following section describes the programming paradigm of the GSM module in detail.

*Implementation of Auto-Messaging System*

The SIM900D is connected to the MCU (Friendly ARM) by USB-Serial cable for receiving instructions from MCU in order to send alerting short messages to the respective official or person in charge. The instructions consist of *commands* and *flow control.*

Command to GSM modules, in general, starts with the prefix *AT* where *AT* stands for "Attention". This prefix must be set at the beginning of each command to the SIM900D at the same time end of each command must be terminated with $<CR>$ or $<nr>$ i.e. the carriage return. Note that these terminating characters can be fed either in equivalence hexadecimal, octal, or decimal form of ASCII values as well. List of essential commands are given in Table 5. The general form of instructing SIM900D from a terminal emulator is *echo < COMMAND > /dev/ttyUSB0*; where COMMAND is the respective AT command set (refer Table 5) and /dev/ttyUSB0 refers the port of the MCU/PC at which the SIM900D is connected. For instance, *echo"ATD0800453947;" > /dev/ttyUSB0* will dial to the mobile whose number is 0800453947 (this is a mobile phone number in Thailand). Note that by using '+' with the respective phone number it is possible to dial internationally with appropriate codes.

Flow control is very important for correct communication between the GSM engine and DTE (Data Terminal Equipment) as there are different rate of receiving and transmitting between the devices involve in communication. There are, basically, two approaches to achieve data flow control: *software flow control* and *hardware flow control*. SIM900D support both the kinds. However, AMSMS employs hardware flow control. The Hardware Flow Control is achieved by controlling the RTS/CTS (Request-to-send/Clear-to-send) line. When the data transfer should be suspended, the CTS line is set inactive until the transfer from the receiving buffer has completed. When the receiving buffer is ready to receive more data, CTS goes active once again. The RTS/CTS lines are present on Friendly ARM platform which hardware flow control enabled.

## IV. Testing

*GSM Module Testing*

The GSM Module (SIM900D) was connected to PC via a USB-Serial cable. Then, the communication terminal- GTKterm was configured as follows; *Port = /dev/ttyUSB0, Speed = 115200, Parity = None, Bits = 8, Stop bits = 1,* and *Flow Control = None*. The following sections describe how to perform operations in GSM module.

*1) Dialing a Phone Number:* After the proper configuration on the terminal, *<AT>* command was sent to check the connection between PC and GSM module. The GSM module responded *OK* i.e. the communication has been successfully established and the GSM module is ready for further operation. Then, *<AT+COPS?>* command was executed to check availability of a network provider at the module. The GSM module acknowledged *TH, GSM* i.e. the there is AIS GSM (one of the Thai mobile network providers) available at the module. The testing further continued with *ATD0895092020* command and the GSM module acknowledged *OK*. After a few seconds of delay, the phone 0895092020 received the call. It was hung up by the command *ATH*. Thus, the testing results showed that the GSM module functioned as desired.

*2) Sending Text Message:* Since the system AMSMS requires the text messaging functionality, testing was further continued to ensure the operation. Initially, text messaging mode of the GSM module was set by *AT+CMGF=1* (refer Table - 5 to understand the command). Note that the commands AT and *AT+COPS?* Were skipped since they were done earlier during dialing test and this testing is the continuation of that. The GSM module acknowledged OK. Consequently, a text message was sent by using *AT+CMGS* command. After a few seconds of delay, the particular text was received by the destined mobile phone.

Table 5: Command List of GSM module [12].

| Command | Description |
|---|---|
| AT | Check a response |
| AT + COPS? | Read a cell operator |
| ATD"PhoneNumber; " | Make a call |
| ATH | Hang up command |
| AT + CMGF =< mode > < CR > | Set messaging mode |
| AT + CMGS = "PhoneNumber" | Set a destination phone number |
| < Message >< CR > n032 | Send a text message |

*3) MCU Controlled Messaging Test:* The testing was conducted by interfacing the GSM module with Mini2440 via USB-Serial connector and instructing the GSM module from GTK terminal which was configured to communicate Mini2440. Figure - 7 shows the connection method between GSM module, Mini2440, and a PC. Then, all the operations tested earlier were tested from the terminal of Mini2440 by using *echo* command. Similar to the previous testing this testing was, also, successful.

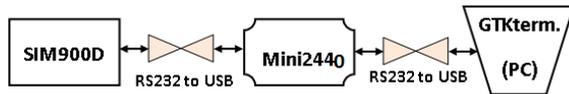

Figure 7: GSM module Interface.

*Operational Testing*

The system has four ADCs for three different observations i.e. detection of smoke, over heat, or trespassing as described in section 3. The aforementioned three conditions are programmed accordingly where a status variable will be updated as shown in the Table 6. In order to test the functionality of the system the sensors were gone through various states by means of the following methods: (i). ADC0 is varied manually (i.e.by rotation), (ii). ADC1 is varied by exposing to variation of heat, (iii). ADC2 is varied manually, and (iii). ADC3 is varied by blocking and exposing to a light source. For every change the output of the particular sensors were tested by a digital multi-meter (DMM). This testing was done over and over and the outcome was checked from the log file. Sample of a log file is shown in Figure – 9.a. Where, the log shows the respective update for each change in the condition. Finally, the overall system was tested for all the conditions. AMSMS generated alarming text message automatically and the message was received by desired mobile phone. A sample of received message at the mobile phone is shown in Figure – 9.b. Note that for overheat detection testing; it is more practical to vary the threshold instead of exposing the sensors to high temperature.

Table 6: The Status Variables and Conditions.

| Case | Status | | |
|---|---|---|---|
| Nothing Detected | 0 | 0 | 0 |
| Smoke detected | 0 | 0 | 1 |
| Overheat detected | 0 | 1 | 0 |
| Smoke and Overheat detected | 0 | 1 | 1 |
| Trespassing Detected | 1 | 0 | 0 |
| Smoke and Trespassing detected | 1 | 0 | 1 |
| Overheat and Trespassing Detected | 1 | 1 | 0 |
| Smoke, Overheat and Trespassing Detected | 1 | 1 | 1 |

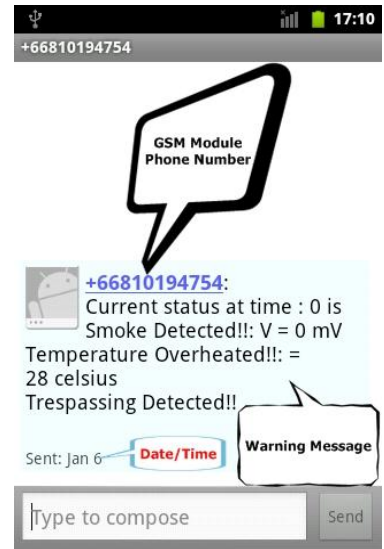

(a)

(b)

Figure 8: Test Results: (a). Content of the Generated Log File, (b) Received Message from AMSMS.

## V. CONCLUSION

The Auto-Monitoring and Short-Messaging-Service System is an ideal low cost surveillance system for a real-time critical operational environment. The entire system is successfully prototyped, its operation has tested and verified for its perfectness. The results showed that the system satisfied the functional requirements. The prototype functionally works well; it can be improved to be a complete system with dedicated circuits, stable connections with more features including multiple MCUs networked as a client-server network for detecting other harmful particles such as chemicals in a real-time operating environment. Hence, the system can be improved to have its own power source rather than relying on external power supply and LCD for interacting with the user. Thus, once the further improvements are carried out to enhance the features of the system it will be a cost effective commercial product, in the future.